%% file: main.tex
\documentclass[conference, a4paper]{IEEEtran}

\usepackage{graphicx}

\usepackage{CJKutf8}

\usepackage{multirow}

\usepackage{amsmath}
\usepackage{amsfonts}

\usepackage{comment}
\usepackage{booktabs}
\usepackage{dirtytalk}
\usepackage{cite}

\usepackage{threeparttable} 

\makeatletter
\let\MYcaption\@makecaption
\makeatother

\usepackage[font=footnotesize]{subcaption}

\makeatletter
\let\@makecaption\MYcaption
\makeatother

\usepackage{subcaption}
\usepackage{float}

\ifCLASSINFOpdf
\else
\fi

\begin{document}

\title{Bulk Production Augmentation Towards Explainable Melanoma Diagnosis}

\makeatletter
\newcommand{\linebreakand}{%
  \end{@IEEEauthorhalign}
  \hfill\mbox{}\par
  \mbox{}\hfill\begin{@IEEEauthorhalign}
}
\makeatother

\author{\IEEEauthorblockN{Kasumi Obi}
\IEEEauthorblockA{Applied Informatics,  Graduate School of\\Science and Engineering\\
Hosei University, Tokyo, Japan\\
Email: kasumi.obi.6e@stu.hosei.ac.jp}
\and
\IEEEauthorblockN{Quan Huu Cap}
\IEEEauthorblockA{Applied Informatics, Graduate School of\\Science and Engineering\\
Hosei University, Tokyo, Japan\\
Email: huu.quan.cap.78@stu.hosei.ac.jp}
\and
\IEEEauthorblockN{Noriko Umegaki-Arao}
\IEEEauthorblockA{Department of Dermatology,\\
Tokyo Women's Medical University\\
Medical Center East, Tokyo, Japan\\
Email: umegaki.noriko@twmu.ac.jp}
\linebreakand 
\IEEEauthorblockN{Masaru Tanaka}
\IEEEauthorblockA{Department of Dermatology,\\
Tokyo Women's Medical University\\
Medical Center East, Tokyo, Japan\\
Email: masarutanaka@1984.jukuin.keio.ac.jp}
\and
\IEEEauthorblockN{Hitoshi Iyatomi}
\IEEEauthorblockA{Applied Informatics, Graduate School of\\Science and Engineering\\
Hosei University, Tokyo, Japan\\
Email: iyatomi@hosei.ac.jp}}

\maketitle

\begin{abstract}
    \input{00_abstract}
\end{abstract}


\IEEEpeerreviewmaketitle

\section{Introduction}
    \input{01_introduction}

\section{Method}
    \input{02_method}

\section{Experiments}
    \input{03_experiments}

\section{Results \& Discussion}
    \input{04_result}

\section{Conclusion}
    \input{05_conclusion}

\bibliographystyle{IEEEtran}
\bibliography{references}

\end{document}

%% file: 00_abstract.tex
Although highly accurate automated diagnostic techniques for melanoma have been reported, the realization of a system capable of providing diagnostic evidence based on medical indices remains an open issue because of difficulties in obtaining reliable training data. 
In this paper, we propose bulk production augmentation (BPA) to generate high-quality, diverse pseudo-skin tumor images with the desired structural malignant features for additional training images from a limited number of labeled images. 
The proposed BPA acts as an effective data augmentation in constructing the feature detector for the atypical pigment network (APN), which is a key structure in melanoma diagnosis. 
Experiments show that training with images generated by our BPA largely boosts the APN detection performance by 20.0 percentage points in the area under the receiver operating characteristic curve, which is 11.5 to 13.7 points higher than that of  conventional CycleGAN-based augmentations in AUC.

%% file: 01_introduction.tex
Melanoma is the most serious type of skin cancer, and it progresses very quickly. 
The WHO reported 287,723 new cases and 60,712 fatalities in 2018 \cite{who}. 
Melanoma is normally diagnosed using dermoscopy (a dermatological microscope). 
However, it requires dermatologist experience, and this sometimes lacks objectivity \cite{argenziano2003dermoscopy}. 
For these reasons, many studies on the development of autonomous melanoma diagnosis systems using dermoscopy images have been proposed.

Prior to the advancement of deep learning techniques, a typical system using classical pattern recognition consists of the following three steps: segmentation of tumor areas \cite{iyatomi2006quantitative,celebi2009lesion}, feature extraction based on medical knowledge, and the final classification \cite{hoffmann2003diagnostic,iyatomi2008improved,ganster2001automated}. 
There have also been some attempts to estimate well-known medical indicators \cite{barhoumi2014pigment, thon2012bayesian, celebi2008automatic, sadeghi2013detection}, such as the ABCD rule \cite{stolz1994abcd} and the 7-point checklist \cite{argenziano1998epiluminescence}.

Despite achieving excellent results, most of these techniques are based on small datasets, so their practicality and generalization performance are still questionable. 
Recently, many studies have reported excellent results, thanks to the rapid progress of deep learning techniques and the release of large-scale dermoscopy image datasets \cite{combalia2019bcn20000,tschandl2018ham10000,codella2018skin}. 
For instance, Gessert et al. proposed a high-accuracy skin lesion diagnosis system \cite{gessert2020skin} that advances ensemble learning of multiple deep learning models. 
Kitada et al. proposed an unlabeled dermoscopy images classifier \cite{kitada2018skin} using semi-supervised learning and body hair augmentation. 

Providing evidence for diagnosis results is very crucial for making medical decisions, such as the diagnosis of cancer. 
Model interpretability has become a more important topic in recent deep learning studies. 
Several studies have proposed explainable melanoma diagnosis systems by utilizing GradCAM \cite{selvaraju2017grad} to provide potential malignant regions \cite{han2018classification} or by improving the intuitiveness of heatmaps to make the model more explainable \cite{zhang2019attention}. 
However, the resulting heatmap is not necessarily interpretable for users because it is a display based on the gradient of the discriminator obtained by training and is not based on medical knowledge.
We believe that providing quantitative diagnosis evidence, such as the 7-point checklist, can improve model interpretability. 

As a related study, Murabayashi et al. \cite{murabayashi2019towards} proposed a method to quantitatively estimate each measure of the 7-point checklist by taking advantage of semi-supervised learning with virtual adversarial training \cite{miyato2018virtual} and multi-task learning \cite{caruana1997multitask}.
They trained their model on only 226 labeled images with the 7-point checklist by four dermatologists and 9,124 images with labels of benign/malignant information. 
As a result, they were able to estimate each criterion on the 7-point checklist with a deviation equivalent to that of a dermatologist, and the diagnostic accuracy was significantly improved by 7.5\%.
However, the quantitative and extensive assessment of these clinical indicators is still limited because of the lack of large-scale images labeled with these indicators \cite{iyatomi2007parameterization, murabayashi2019towards}.
Labeling large datasets with multiple dermatologists is expensive, and obtaining such datasets is difficult. 
Thus, addressing the lack of dataset could improve model performance. 

Recently, models based on generative adversarial networks (GANs) \cite{goodfellow2014generative} have been successfully used in various fields. 
In the application for computer vision, progressive growing GAN (PGGAN) \cite{karras2017progressive} achieved the synthesis of high-resolution realistic images by progressively enlarging image resolution during the training process. 
CycleGAN\cite{zhu2017unpaired} learns the relationship between images from two different domains to translate styles without paired training images; this process has been applied in a variety of fields. 

Some studies have reported improving medical diagnosis accuracy by increasing the number of training images using GAN-based image generation models with a limited number of training data \cite{frid2018gan,han2020infinite}.
However, since GAN learns to estimate the distribution of training data, it cannot expand the distribution itself because the generated image is basically an interpolated sample from the dataset \cite{salimans2016improved}. 
This also applies to CycleGAN;  the variety of generated images is limited because CycleGAN can only perform one-to-one conversion.

Therefore, in this paper, we propose bulk production augmentation (BPA), which is a framework for generating a large volume and variety of pseudo-images of the desired category, and we apply it to the quantification task of the dermoscopic structure toward an explainable melanoma diagnosis.
The proposed BPA can be divided into two phases: (1) bulk production phase and (2) feature transition phase. 
In the bulk production phase, PGGAN augments high-resolution images of the nevus, which are widely available and have less malignant features.
They are used as the foundation for adding malignant features in the second phase. 
Then, in the feature transition phase, malignant features are added to the generated nevus images using CycleGAN. 
With these steps, the system can generate a large number of varied dermoscopy images with malignant features.

In the evaluation experiments, we tested the effectiveness of the proposed BPA framework on the atypical pigment network (APN), an indicator registered as a major criterion in the 7-point checklist. 
For the training images, we used 230 images labeled as APN and 10,000 nevus images from public datasets. 

With respect to the evaluation of our proposed data generation (i.e., augmentation) framework, direct evaluation by dermatologists is extremely expensive and infeasible. 
We therefore trained two types of classifiers to confirm if we could improve classification performance by augmenting the data with our framework. 
The first type of classifier is the detectors of APN structures, and we used them to verify how data augmentation by BPA improves detection performance. 
The second type is a melanoma-nevus classifier, which is used to verify whether the images generated by the  APN's constructional features by BPA improves the grade of malignancy compared with that before the assignment.

\input{images/tex/7-point}

%% file: images/tex/7-point.tex
\begin{table}[]
\centering
\caption{7-point checklist criteria}
\label{tab:7-point}
\begin{tabular}{@{}lc@{}}
\toprule
Major criteria               & weight               \\ \midrule
1. Atypical pigment network          & $\times$2                   \\ \midrule
2. Blue-whitish veil         & $\times$2                   \\ \midrule
3. Atypical vascular pattern & $\times$2                   \\ \midrule
Minor criteria               & \multicolumn{1}{l}{} \\ \midrule
4. Irregular streaks         & $\times$1                   \\ \midrule
5. Irregular pigmentation    & $\times$1                   \\ \midrule
6. Irregular dot/globules    & $\times$1                   \\ \midrule
7. Regression structures     & $\times$1                   \\ \bottomrule
\end{tabular}
\end{table}

%% file: 02_method.tex
\subsection{The 7-point checklist - preparation}

\input{images/tex/apn}
\input{images/tex/dataset}

The 7-point checklist is a well-known diagnostic method using dermoscopy. 
This checklist requires the identification of seven dermoscopic structures, as shown in Table \ref{tab:7-point}. 
The first three are categorized as major criteria, and the latter four are minor criteria. 
The score for a skin lesion is determined as the weighted sum of the structures present in it. 
Using the 7-point checklist, the total score (TS) is calculated as
\begin{equation}
    \mathrm{TS} = \mathrm{\#major} \times 2 + \mathrm{\#minor},
\end{equation}
where \#major and \#minor are the number of major and minor dermoscopic structures present in the image, respectively. Accordingly, the TS ranges between 0 and 10. 
If the TS is greater than or equal to 3, the lesion is considered malignant. 
According to the literature, diagnostic accuracy based on the 7-point checklist by 40 experts is 75.0\% in sensitivity and 76.2\% in specificity \cite{argenziano2003dermoscopy}. 
In this paper, we focus on the classification of the APN as one of the major criteria.
Fig. \ref{fig:apn_sample} shows a typical example of dermoscopy images with APN structures. 
While the APN is a strong indicator of melanoma diagnosis, it should be noted that the presence of this structure does not necessarily indicate malignancy. 

\input{images/tex/architecture}

\subsection{Bulk production augmentation (BPA)}
In this paper, we propose the bulk production augmentation framework (BPA) in order to develop an interpretable automated melanoma diagnosis system. 
The proposed BPA is a novel and effective data augmentation framework when there are very little supervised data for training. 
Our BPA is composed of two phases: (1) the bulk production phase and (2) the feature transition phase.
Fig. \ref{fig:archi} shows the flow of our proposed BPA framework. 
Each phase uses different kinds of GANs. 
The first phase involves the generation of a large number of nevus images, and the second phase involves the transfer of APN features to the diverse nevus images generated in the first phase. 

\subsubsection{The bulk production phase}
The bulk production phase generates a number of diverse images that serve as the basis for the characterization in the second phase. 
In this paper, a nevus (i.e., a benign pigmented skin tumor) is used as the base image. 
A nevus has a low proportion of malignant dermoscopic features and is readily available in large quantities.  
It also has attractive attributes as a basis for transferring disease features. 
In this phase, based on the nevus images, a large number of pseudo-nevus images are generated using an adversarial generation network to further increase diversity. 
We used PGGAN \cite{karras2017progressive} for image generation. 
PGGAN takes a latent vector (noise) as an input and gradually scales up the image size, eventually generating various high-quality images.

\subsubsection{The feature transition phase}
The feature transition phase assigns the desired features to the images generated in the previous bulk production phase. 
Specifically, in this paper, CycleGAN \cite{zhu2017unpaired} is used to assign APN features to the nevus training datasets (see Table~\ref{tab:datasets}). 
CycleGAN learns the data distribution of two domains to achieve a style transformation without the need for image pairs. 

The proposed BPA solves the problem of quantity and diversity of training images. The bulk generation stage is the cornerstone of this proposal's data augmentation, as it allows for the generation of virtually unlimited base images.

%% file: images/tex/apn.tex
\begin{figure}[t]
    \centering
    \includegraphics[width=0.6\linewidth]{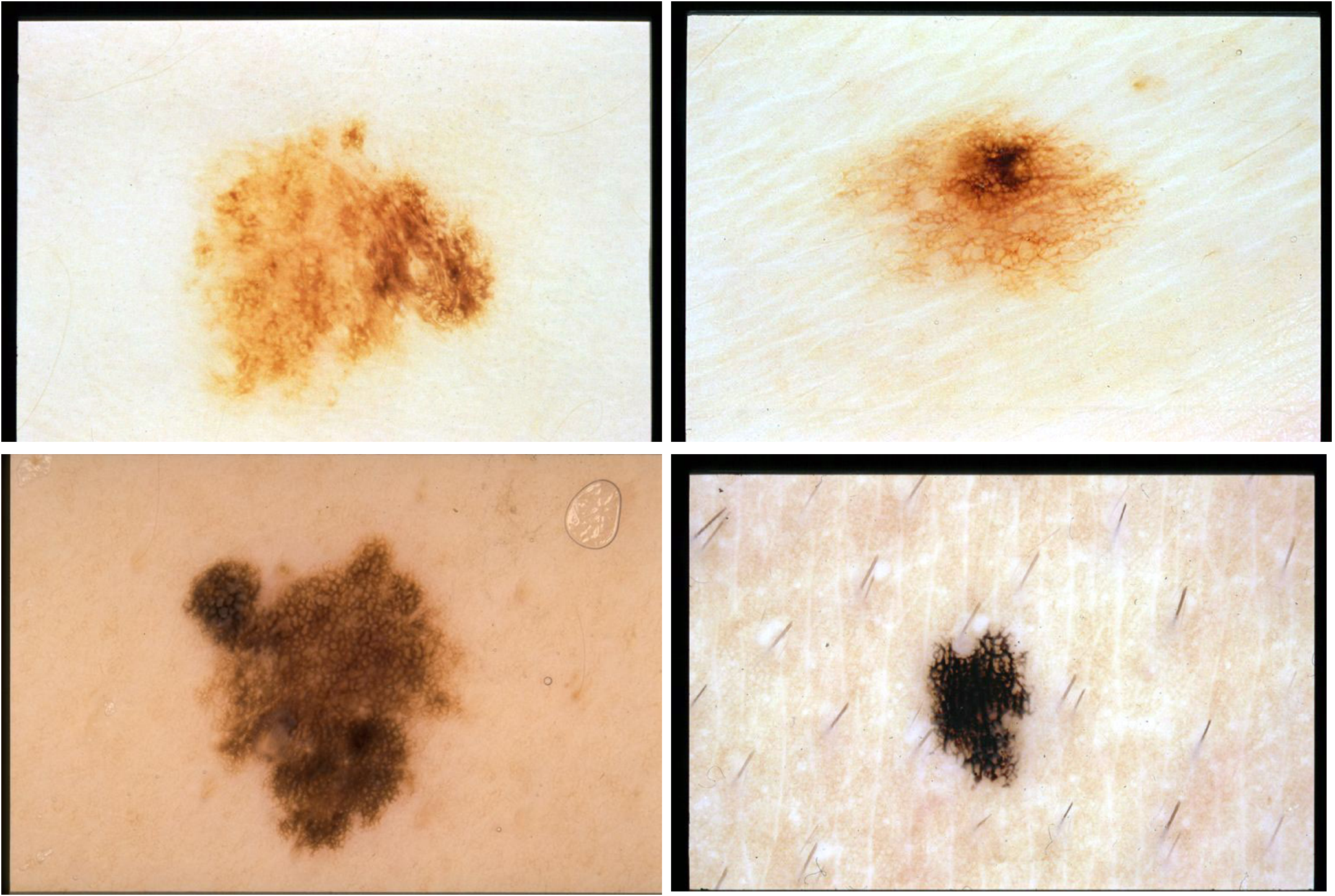}
    \caption{
    Dermoscopy images with atypical pigment network (APN) structures.
    }
    \label{fig:apn_sample}
\end{figure}

%% file: images/tex/dataset.tex
\begin{table*}[]
\centering
\caption{Training datasets for apn feature detectors to evaluate the proposed bpa}
\label{tab:datasets}
\begin{threeparttable}
\begin{tabular}{@{}llrrrrr@{}}
\toprule
  & Dataset         & \multicolumn{1}{l}{nevus} & \multicolumn{1}{l}{nevusG} & \multicolumn{1}{l}{APN} & \multicolumn{1}{l}{APN\_nevus} & \multicolumn{1}{l}{APN\_nevusG} \\ \midrule
(A) & baseline          & 10,000                         &                             & 230                          &                                  &                                    \\
(B) & CycleGAN          & 10,000                         &                             & 230                          & 10,000                           &                                    \\
(C) & simplified BPA & 10,000                         &                             & 230                          &                                  & 10,000                             \\
(D) & (proposed) BPA    & 10,000                         & $^{\dagger}$10,000          & 230                          &                                  & $^{\dagger}$20,000                 \\ \bottomrule
\end{tabular}
\begin{tablenotes}
\item[$\dagger$] Thanks to the bulk production phase of the proposed BPA, more images can be
generated than the original number of data and can be added to the training. 
\end{tablenotes}
\end{threeparttable}
\end{table*}

%% file: images/tex/architecture.tex
\begin{figure}[t]
    \centering
    \includegraphics[width=0.8\linewidth]{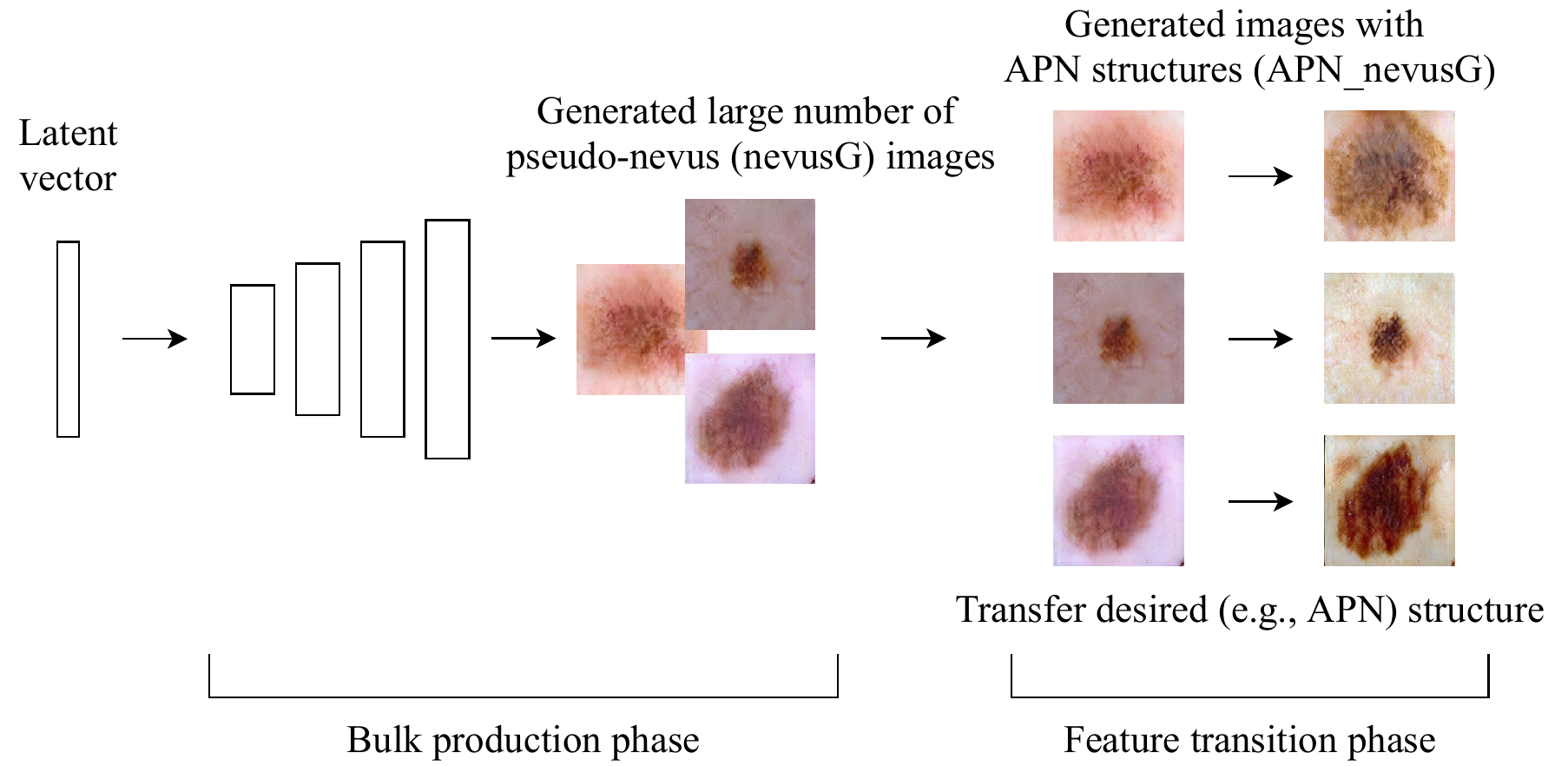}
    \caption{
    The overview of the proposed bulk production augmentation (BPA).
    }
    \label{fig:archi}
\end{figure}

%% file: 03_experiments.tex
In this paper, we used the ISIC 2019 open dermoscopy image dataset~\cite{tschandl2018ham10000,codella2018skin,combalia2019bcn20000} and an independent dataset presented in the literature \cite{kawahara2018seven}. 
Specifically, the ISIC 2019 dataset consists of 12,875 nevi and 4,522 melanomas. 
The latter dataset contains 230 images that were identified as having APN features. 
For our experiments, the resolution of images to be used in both phases of the BPA was set to 256$\times$256 pixels.

\subsection{Training the BPA}
For the training of PGGAN in the bulk production phase, a total of 6,816 images of nevi without artifacts, such as hairs, measures, and pen writings, were selected from the ISIC 2019 dataset to generate high-quality skin tumor images (i.e., pseudo-nevus images).

For the training data of CycleGAN in the feature transition phase, we used the aforementioned 230 dermoscopy images with APN features and 2,000 nevi (randomly collected) from the ISIC 2019 dataset, considering the balance in the number of data. 

We followed the default settings from both PGGAN and CycleGAN to train our models. 
Please refer to the original literature for more training details \cite{karras2017progressive,zhu2017unpaired}. 

\input{images/tex/APN_Gnevus_sample}

\subsection{Evaluation the BPA}
The diagnosis of APN features is known to vary widely even among skilled dermatologists, as mentioned in \cite{argenziano2003dermoscopy}. 
Therefore, the  of the generation of dermoscopy images requires agreement on the diagnosis of multiple dermatologists. 
However, this is impractical for a large amount of output results.
We thus decided to perform the following two indirect but quantitative evaluations (1) Effect on APN detection; (2) Effect on malignancy grading.

\subsubsection{Effect on APN detection}
To evaluate the effect of the proposed BPA framework, deep neural network classifiers for estimating the presence or absence of APN features were trained on several groups of training data, and their performance was compared.
As for the evaluation dataset, 134 and 100 dermoscopy images with and without APNs, respectively, were selected from the ISIC 2019 dataset based on the agreement of two skilled dermatologists. 
Note here that we excluded pigmented skin lesions in acral volar regions because they have completely different dermoscopic structures  \cite{saida2004significance, iyatomi2008computer}.
The APN classifiers were fine-tuned based on the EfficientNet-B1 network \cite{tan2019efficientnet}, which has been reported to have excellent classification performance.
A comparison of the conditions of the training data, including the proposed BPA, is shown in Table \ref{tab:datasets}.

For the sake of brevity, the following dataset notation will be used thereafter:
\begin{itemize}
    \item \textbf{nevus}: Real nevus images (original training data)
    \item \textbf{nevusG}: Generated pseudo-nevus images from the bulk production phase (used as base skin tumor images in the proposed BPA)
    \item \textbf{APN}: Real images identified with APN structures (original training data)
    \item \textbf{APN\_nevus}: Generated images with APN structures by CycleGAN from real nevus images
    \item \textbf{APN\_nevusG}: Generated images with APN structures by CycleGAN from generated nevus images (nevusG)
\end{itemize}
Note here that \say{APN\_nevusG} is the main outcome of our proposed BPA. 

The conditions (i.e., training datasets for the APN classifiers) can be summarized as follows:
\begin{itemize}
    \item[(A)] \textbf{baseline}: We used 10,000 real nevus images and 230 real APN images.
    \item[(B)] \textbf{CycleGAN}: In addition to the baseline, 10,000 generated APN images from nevus images (APN\_nevus) are added to the training.
    \item[(C)] \textbf{simplified BPA}: In addition to the baseline, 10,000 generated APN images from nevusG (APN\_nevusG) are added to the training.
    \item[(D)] \textbf{(proposed) BPA}: In addition to the baseline, 10,000 generated nevus images (nevusG) and 20,000 APN images (APN\_nevusG) are added to the training.
\end{itemize}
The condition (C) is a simplified version of BPA that we introduced to compare the effect of data augmentation by BPA with that of the conventional CycleGAN method. 
With this condition, we focused on evaluating the efficiency of generating diverse images by the bulk production phase.

\subsubsection{Effect on malignancy grading}
We evaluated the validity of the generated APN feature-assigned images from a different perspective. 
Because APN features are an indicator of a melanoma diagnosis, an improvement in malignancy should be expected if APN features are properly assigned by BPA.
In this experiment, we built a deep classifier capable of classifying either melanoma or nevus with high accuracy, and we evaluated whether the classifier increases the detection degree of malignancy by assigning APN features. 
To this end, we used the EfficientNet-B1\cite{tan2019efficientnet} network as the backbone of our grading classifier. 
The model was trained on 8,000 nevi and 4,000 melanomas from the ISIC dataset, and we confirmed that it achieved a sensitivity of 88.8\% and a specificity of 94.2\% (F1-score: 0.914) for 500 test cases each of nevus and melanoma. 

Our classifier used the sigmoid function as its output, so the malignancy grading score ranges from 0.0 to 1.0 in our experiment. 
We then compared the grading performance of the trained classifier on the following five datasets: real nevus, generated nevus (nevusG), generated APN images from the real nevus (APN\_nevus), generated APN images from the generated nevus (APN\_nevusG), and real APN images.

\subsubsection{Training details of the classifiers}
All the above classifiers for evaluating our BPA were fined-tuned from the pre-trained EfficientNet-B1 model. 
The input images were resized to 240$\times$240 pixels, which was optimally designed for EfficientNet-B1. 
Random resize cropping and random horizontal flip data augmentation were performed with a scale in the range of 0.5 to 1.0 during the training. 
Furthermore, RandAugment\cite{cubuk2020randaugment}, a state-of-the-art data augmentation method, was applied to all models with $N=6$ and $M=8$ based on the results of preliminary experiments. 
Since the datasets in Table \ref{tab:datasets} are extremely imbalanced, we introduced the weighted loss \cite{huang2016learning} to address the class-imbalanced problem. 
The momentumSGD \cite{qian1999momentum} optimizer with a learning rate of $1 \times 10^{-5}$ and weight decay for each epoch of $1 \times 10^{-6}$ are used to train all classifiers. 
For more details of the configurations and training, please refer to the original EfficientNet paper~\cite{tan2019efficientnet}.

%% file: images/tex/APN_Gnevus_sample.tex
\begin{figure*}[]
    \centering
    \includegraphics[width=0.6\linewidth]{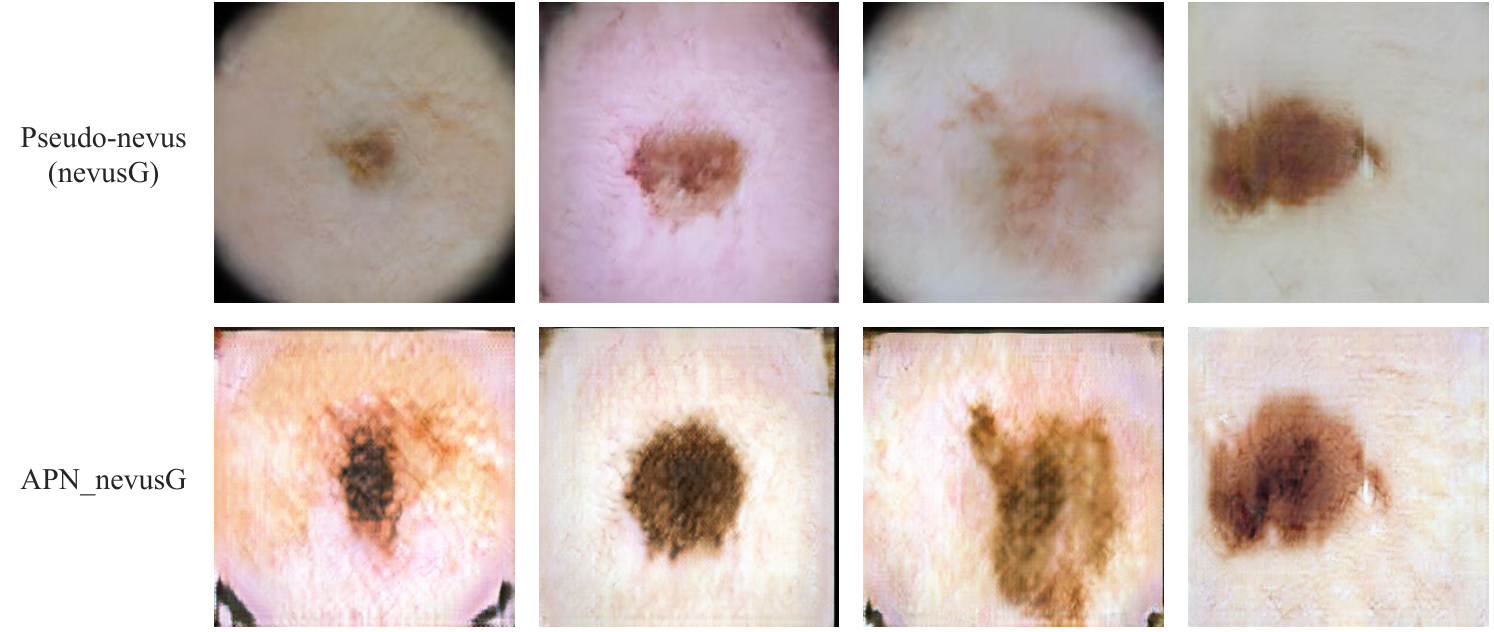}
    \caption{
    Example of generated dermoscopy images. The upper images are generated nevus images by the bulk production phase (nevusG). 
    The lower images are transformed with the APN structure from the images by the feature transition phase (APN\_nevusG). 
    }
    \label{fig:apn_gnevus_samples}
\end{figure*}

%% file: 04_result.tex
\input{images/tex/result}
\input{images/tex/roc}

\subsection{Improving APN feature detection with BPA}
Fig. \ref{fig:apn_gnevus_samples} shows the example images  generated by the proposed BPA.
The generated images are visually natural.
A summary of the detection performance for APN features is shown in Table \ref{tab:result}, and their receiver operating characteristic (ROC) curves are shown in Fig. \ref{fig:roc_curve}. 

For the baseline (A), with a small number of real APN images, the state-of-the-art EfficientNet is unable to detect the APN structure despite the introduction of RandAugment data augmentation and the weighted loss. 
Two CycleGAN-based models (B and C) improve the APN detection performance by 6.3 and 8.5 points in the area under the ROC curve (AUC), respectively, whereas the proposed BPA (D) shows a significant improvement by 20.0 points (0.722 in the AUC). 
With the introduction of CycleGAN (B), there is an improvement in accuracy, but the accuracy obtained is insufficient. 
The result from (C), the simplified version of the proposed BPA, which assigns APN features to the pseudo-nevus images (nevusG), shows a slight improvement from (B). 
This may be due to the generated images being different from any of the original images, making them more effective in the data augmentation process. 
In the proposed framework (D), BPA successfully generates a large number of diverse and high-quality images with APN structures (APN\_nevusG) based on a large number of \textit{base} pseudo-nevus images (nevusG) generated in the bulk production phase. 
These BPA features are the reason for the significant improvement in APN detection. 

\subsection{Assessing the impact of APN features on grading malignancy}
Fig. \ref{fig:density} shows the distribution of malignancy in each image by the melanoma-nevus discriminator. 
A value close to 1 indicates a higher grade of malignancy. 
While the result on the real APNs has score peaks at 0 and 1, the other results only have large peaks at 0. 
This is because a certain number of nevi have APN features. 

The score distribution of the generated APN\_nevusG images by our BPA (yellow) shows higher malignancy than that of the real nevus images (red) and nevusG images (green). 
It (yellow) shows a higher probability than APN\_nevus (blue) in both areas that are close to grades 0 (benign) and 1 (melanoma), confirming that the APN features are appropriately assigned. 

Overall, the proposed BPA can assign malignancy characteristics at about the same level as when the feature translation is performed from real data. 
The validity of the proposed framework is also confirmed by the fact that the malignancy distribution of the proposal (yellow) is closest to the distribution with real APN features (purple). 
Here, we need to emphasize that our proposed framework differs from conventional methods because it can generate a large number of images with the desired features, which can be a useful data augmentation method. 
This advantage led to the significant improvement in the detection of APN features in our experiments. 
This supports the fact that the above (B) is less effective than our proposed (C) and (D). 

\input{images/tex/density}

%% file: images/tex/result.tex
\begin{table*}[]
\centering
\caption{Comparison of the detection performance for atypical pigment network (APN) structures}
\label{tab:result}
\begin{tabular}{@{}lrrrrr@{}}
\toprule
\multicolumn{1}{c}{Dataset} & \multicolumn{1}{c}{Accuracy (\%)} & \multicolumn{1}{c}{Recall (\%)} & \multicolumn{1}{c}{Precision (\%)} & \multicolumn{1}{c}{F1 (\%)} & \multicolumn{1}{c}{AUC} \\ \midrule
(A) baseline              & 53.8                        & 69.4                           & 58.1                              & 63.3                       & 0.522                   \\
(B) CycleGAN                  & 59.8                        & 64.9                           & 64.9                              & 64.9                       & 0.585                   \\
(C) simplified BPA       & 58.5                        & 54.5                           & 70.0                              & 60.1                       & 0.607                   \\
\textbf{(D) (proposed) BPA}   & \textbf{69.7}               & \textbf{77.6}                  & \textbf{71.7}                     & \textbf{74.6}              & \textbf{0.722}          \\ \bottomrule
\end{tabular}
\end{table*}

%% file: images/tex/roc.tex
\begin{figure}[]
    \centering
    \includegraphics[width=0.7\linewidth]{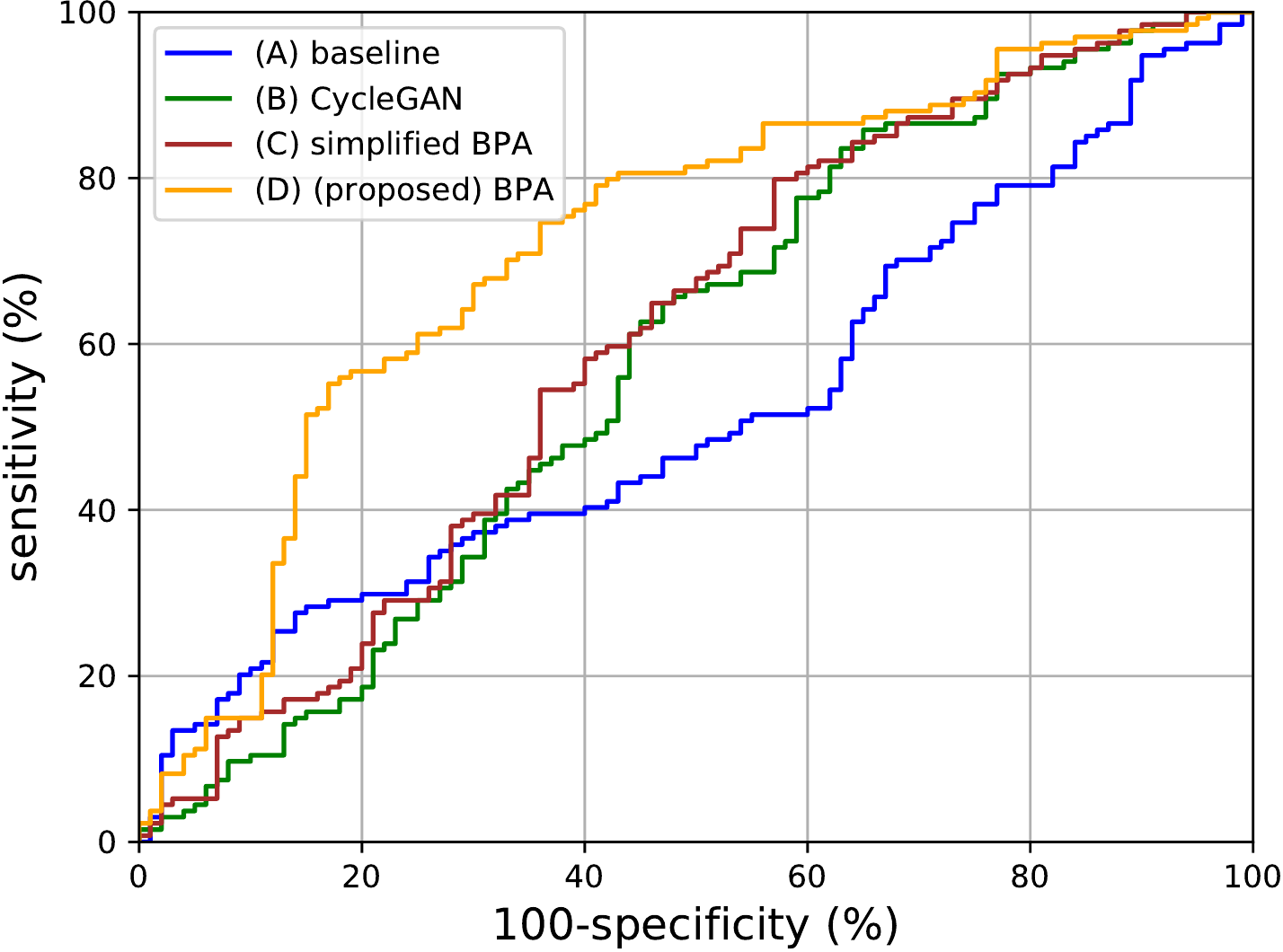}
    \caption{
    Receiver operating characteristic curve for detecting the APN structure. Our proposed bulk production augmentation (BPA) largely improved detection performance. 
    }
    \label{fig:roc_curve}
\end{figure}

%% file: images/tex/density.tex
\begin{figure}[]
    \centering
    \includegraphics[width=0.7\linewidth]{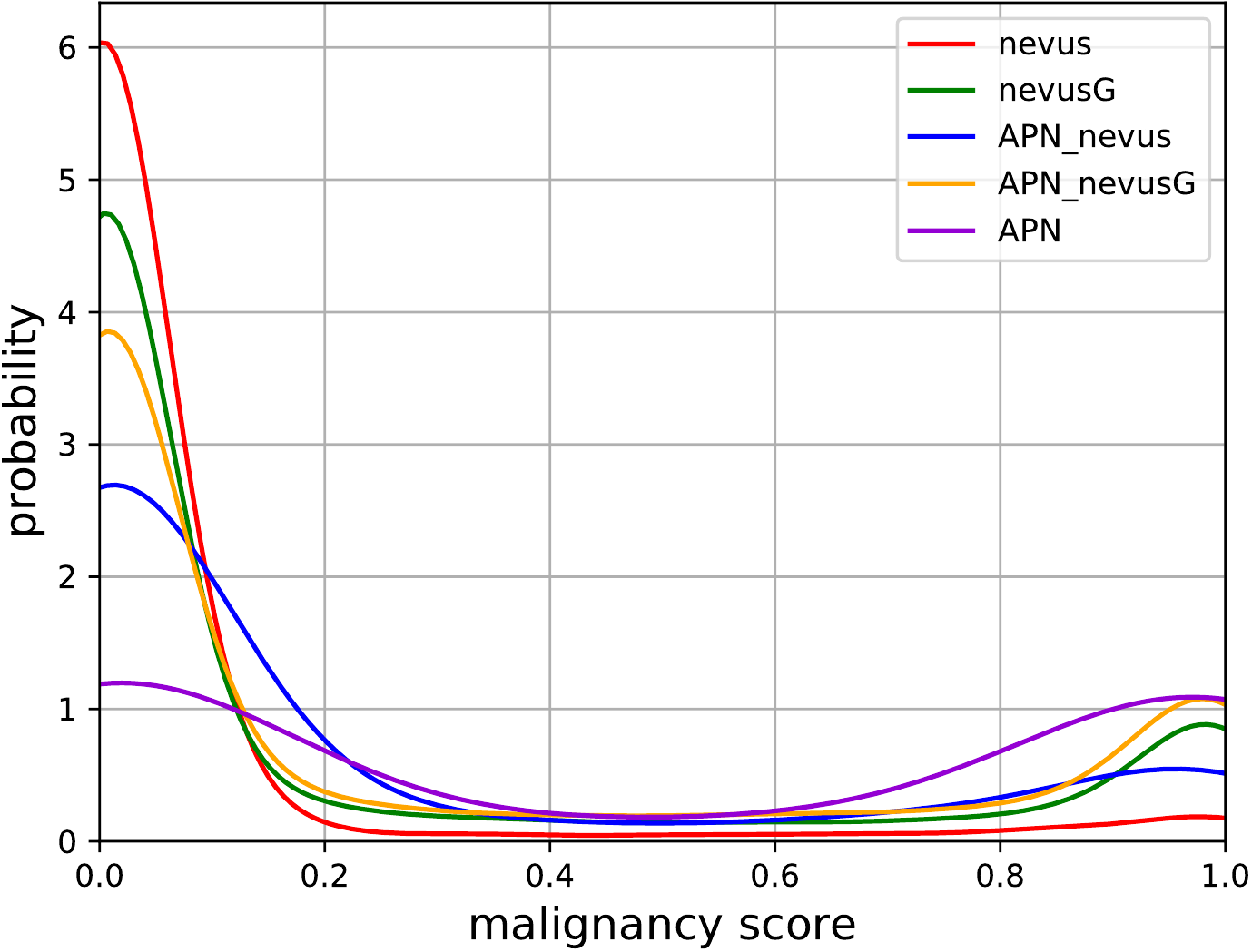}
    \caption{
    Comparison of malignancy score distributions. The images obtained by the proposed BPA show a malignancy distribution more similar to that of the APN.
    }
    \label{fig:density}
\end{figure}

%% file: 05_conclusion.tex
In this paper, we proposed the bulk production augmentation (BPA) for the realization of an automated diagnosis system that can present diagnostic evidence based on medical indices. 
BPA first produces a large number of data with general features to be generated and then transfers the desired features from the previous diverse generated base images. 
BPA is a novel and effective data augmentation framework for improving the interpretability of automated medical diagnosis systems when there are very little supervised data for training. 
We believe that our BPA can be applied for augmenting other dermoscopic structures data defined in the 7-point checklist or other criteria.

%% file: main.bbl
\begin{thebibliography}{10}
\providecommand{\url}[1]{#1}
\csname url@samestyle\endcsname
\providecommand{\newblock}{\relax}
\providecommand{\bibinfo}[2]{#2}
\providecommand{\BIBentrySTDinterwordspacing}{\spaceskip=0pt\relax}
\providecommand{\BIBentryALTinterwordstretchfactor}{4}
\providecommand{\BIBentryALTinterwordspacing}{\spaceskip=\fontdimen2\font plus
\BIBentryALTinterwordstretchfactor\fontdimen3\font minus
  \fontdimen4\font\relax}
\providecommand{\BIBforeignlanguage}[2]{{%
\expandafter\ifx\csname l@#1\endcsname\relax
\typeout{** WARNING: IEEEtran.bst: No hyphenation pattern has been}%
\typeout{** loaded for the language `#1'. Using the pattern for}%
\typeout{** the default language instead.}%
\else
\language=\csname l@#1\endcsname
\fi
#2}}
\providecommand{\BIBdecl}{\relax}
\BIBdecl

\bibitem{who}
\BIBentryALTinterwordspacing
WHO, ``Global cancer observatory,'' \emph{World Health Organization}. [Online].
  Available: \url{https://gco.iarc.fr/}
\BIBentrySTDinterwordspacing

\bibitem{argenziano2003dermoscopy}
G.~Argenziano, H.~P. Soyer, S.~Chimenti, R.~Talamini, R.~Corona, F.~Sera,
  M.~Binder, L.~Cerroni, G.~De~Rosa, G.~Ferrara \emph{et~al.}, ``Dermoscopy of
  pigmented skin lesions: results of a consensus meeting via the internet,''
  \emph{Journal of the American Academy of Dermatology}, vol.~48, no.~5, pp.
  679--693, 2003.

\bibitem{iyatomi2006quantitative}
H.~Iyatomi, H.~Oka, M.~Saito, A.~Miyake, M.~Kimoto, J.~Yamagami, S.~Kobayashi,
  A.~Tanikawa, M.~Hagiwara, K.~Ogawa \emph{et~al.}, ``Quantitative assessment
  of tumour extraction from dermoscopy images and evaluation of computer-based
  extraction methods for an automatic melanoma diagnostic system,''
  \emph{Melanoma Research}, vol.~16, no.~2, pp. 183--190, 2006.

\bibitem{celebi2009lesion}
M.~E. Celebi, H.~Iyatomi, G.~Schaefer, and W.~V. Stoecker, ``Lesion border
  detection in dermoscopy images,'' \emph{Computerized Medical Imaging and
  Graphics}, vol.~33, no.~2, pp. 148--153, 2009.

\bibitem{hoffmann2003diagnostic}
K.~Hoffmann, T.~Gambichler, A.~Rick, M.~Kreutz, M.~Anschuetz,
  T.~Gr{\"u}nendick, A.~Orlikov, S.~Gehlen, R.~Perotti, L.~Andreassi
  \emph{et~al.}, ``Diagnostic and neural analysis of skin cancer (danaos). a
  multicentre study for collection and computer-aided analysis of data from
  pigmented skin lesions using digital dermoscopy,'' \emph{British Journal of
  Dermatology}, vol. 149, no.~4, pp. 801--809, 2003.

\bibitem{iyatomi2008improved}
H.~Iyatomi, H.~Oka, M.~E. Celebi, M.~Hashimoto, M.~Hagiwara, M.~Tanaka, and
  K.~Ogawa, ``An improved internet-based melanoma screening system with
  dermatologist-like tumor area extraction algorithm,'' \emph{Computerized
  Medical Imaging and Graphics}, vol.~32, no.~7, pp. 566--579, 2008.

\bibitem{ganster2001automated}
H.~Ganster, P.~Pinz, R.~Rohrer, E.~Wildling, M.~Binder, and H.~Kittler,
  ``Automated melanoma recognition,'' \emph{IEEE Transactions on Medical
  Imaging}, vol.~20, no.~3, pp. 233--239, 2001.

\bibitem{barhoumi2014pigment}
W.~Barhoumi and A.~Ba{\^a}zaoui, ``Pigment network detection in dermatoscopic
  images for melanoma diagnosis,'' \emph{IRBM}, vol.~35, no.~3, pp. 128--138,
  2014.

\bibitem{thon2012bayesian}
K.~Thon, H.~Rue, S.~O. Skr{\o}vseth, and F.~Godtliebsen, ``Bayesian multiscale
  analysis of images modeled as gaussian markov random fields,''
  \emph{Computational Statistics \& Data Analysis}, vol.~56, no.~1, pp. 49--61,
  2012.

\bibitem{celebi2008automatic}
M.~E. Celebi, H.~Iyatomi, W.~V. Stoecker, R.~H. Moss, H.~S. Rabinovitz,
  G.~Argenziano, and H.~P. Soyer, ``Automatic detection of blue-white veil and
  related structures in dermoscopy images,'' \emph{Computerized Medical Imaging
  and Graphics}, vol.~32, no.~8, pp. 670--677, 2008.

\bibitem{sadeghi2013detection}
M.~Sadeghi, T.~K. Lee, D.~McLean, H.~Lui, and M.~S. Atkins, ``Detection and
  analysis of irregular streaks in dermoscopic images of skin lesions,''
  \emph{IEEE Transactions on Medical Imaging}, vol.~32, no.~5, pp. 849--861,
  2013.

\bibitem{stolz1994abcd}
W.~Stolz, ``Abcd rule of dermatoscopy: a new practical method for early
  recognition of malignant melanoma,'' \emph{European Journal of Dermatology},
  vol.~4, pp. 521--527, 1994.

\bibitem{argenziano1998epiluminescence}
G.~Argenziano, G.~Fabbrocini, P.~Carli, V.~De~Giorgi, E.~Sammarco, and
  M.~Delfino, ``Epiluminescence microscopy for the diagnosis of doubtful
  melanocytic skin lesions: comparison of the abcd rule of dermatoscopy and a
  new 7-point checklist based on pattern analysis,'' \emph{Archives of
  Dermatology}, vol. 134, no.~12, pp. 1563--1570, 1998.

\bibitem{combalia2019bcn20000}
M.~Combalia, N.~C. Codella, V.~Rotemberg, B.~Helba, V.~Vilaplana, O.~Reiter,
  C.~Carrera, A.~Barreiro, A.~C. Halpern, S.~Puig \emph{et~al.}, ``Bcn20000:
  Dermoscopic lesions in the wild,'' \emph{CoRR, abs/1908.02288}, 2019.

\bibitem{tschandl2018ham10000}
P.~Tschandl, C.~Rosendahl, and H.~Kittler, ``The ham10000 dataset, a large
  collection of multi-source dermatoscopic images of common pigmented skin
  lesions,'' \emph{Scientific Data}, vol.~5, p. 180161, 2018.

\bibitem{codella2018skin}
N.~C. Codella, D.~Gutman, M.~E. Celebi, B.~Helba, M.~A. Marchetti, S.~W. Dusza,
  A.~Kalloo, K.~Liopyris, N.~Mishra, H.~Kittler \emph{et~al.}, ``Skin lesion
  analysis toward melanoma detection: A challenge at the 2017 international
  symposium on biomedical imaging (isbi), hosted by the international skin
  imaging collaboration (isic),'' \emph{IEEE International Symposium on
  Biomedical Imaging}, pp. 168--172, 2018.

\bibitem{gessert2020skin}
N.~Gessert, M.~Nielsen, M.~Shaikh, R.~Werner, and A.~Schlaefer, ``Skin lesion
  classification using ensembles of multi-resolution efficientnets with meta
  data,'' \emph{MethodsX}, p. 100864, 2020.

\bibitem{kitada2018skin}
S.~Kitada and H.~Iyatomi, ``Skin lesion classification with ensemble of
  squeeze-and-excitation networks and semi-supervised learning,'' \emph{CoRR,
  abs/1809.02568}, 2018.

\bibitem{selvaraju2017grad}
R.~R. Selvaraju, M.~Cogswell, A.~Das, R.~Vedantam, D.~Parikh, and D.~Batra,
  ``Grad-cam: Visual explanations from deep networks via gradient-based
  localization,'' \emph{IEEE International Conference on Computer Vision}, pp.
  618--626, 2017.

\bibitem{han2018classification}
S.~S. Han, M.~S. Kim, W.~Lim, G.~H. Park, I.~Park, and S.~E. Chang,
  ``Classification of the clinical images for benign and malignant cutaneous
  tumors using a deep learning algorithm,'' \emph{Journal of Investigative
  Dermatology}, vol. 138, no.~7, pp. 1529--1538, 2018.

\bibitem{zhang2019attention}
J.~Zhang, Y.~Xie, Y.~Xia, and C.~Shen, ``Attention residual learning for skin
  lesion classification,'' \emph{IEEE Transactions on Medical Imaging},
  vol.~38, no.~9, pp. 2092--2103, 2019.

\bibitem{murabayashi2019towards}
S.~Murabayashi and H.~Iyatomi, ``Towards explainable melanoma diagnosis:
  Prediction of clinical indicators using semi-supervised and multi-task
  learning,'' \emph{IEEE International Conference on Big Data}, pp. 4853--4857,
  2019.

\bibitem{miyato2018virtual}
T.~Miyato, S.~I. Maeda, M.~Koyama, and S.~Ishii, ``Virtual adversarial
  training: a regularization method for supervised and semi-supervised
  learning,'' \emph{IEEE Transactions on Pattern Analysis and Machine
  Intelligence}, vol.~41, no.~8, pp. 1979--1993, 2018.

\bibitem{caruana1997multitask}
R.~Caruana, ``Multitask learning,'' \emph{Machine Learning}, vol.~28, no.~1,
  pp. 41--75, 1997.

\bibitem{iyatomi2007parameterization}
H.~Iyatomi, H.~Oka, M.~E. Celebi, M.~Tanaka, and K.~Ogawa, ``Parameterization
  of dermoscopic findings for the internet-based melanoma screening system,''
  \emph{IEEE Symposium on Computational Intelligence in Image and Signal
  Processing}, pp. 189--193, 2007.

\bibitem{goodfellow2014generative}
I.~Goodfellow, J.~Pouget-Abadie, M.~Mirza, B.~Xu, D.~Warde-Farley, S.~Ozair,
  A.~Courville, and Y.~Bengio, ``Generative adversarial nets,'' \emph{Advances
  in Neural Information Processing Systems}, pp. 2672--2680, 2014.

\bibitem{karras2017progressive}
T.~Karras, T.~Aila, S.~Laine, and J.~Lehtinen, ``Progressive growing of gans
  for improved quality, stability, and variation,'' \emph{International
  Conference on Learning Representations}, pp. 1--26, 2017.

\bibitem{zhu2017unpaired}
J.-Y. Zhu, T.~Park, P.~Isola, and A.~A. Efros, ``Unpaired image-to-image
  translation using cycle-consistent adversarial networks,'' \emph{IEEE
  International Conference on Computer Vision}, pp. 2223--2232, 2017.

\bibitem{frid2018gan}
M.~Frid-Adar, I.~Diamant, E.~Klang, M.~Amitai, J.~Goldberger, and H.~Greenspan,
  ``Gan-based synthetic medical image augmentation for increased cnn
  performance in liver lesion classification,'' \emph{Neurocomputing}, vol.
  321, pp. 321--331, 2018.

\bibitem{han2020infinite}
C.~Han, L.~Rundo, R.~Araki, Y.~Furukawa, G.~Mauri, H.~Nakayama, and H.~Hayashi,
  ``Infinite brain mr images: Pggan-based data augmentation for tumor
  detection,'' \emph{Neural Approaches to Dynamics of Signal Exchanges}, pp.
  291--303, 2020.

\bibitem{salimans2016improved}
T.~Salimans, I.~Goodfellow, W.~Zaremba, V.~Cheung, A.~Radford, and X.~Chen,
  ``Improved techniques for training gans,'' \emph{Advances in Neural
  Information Processing Systems}, pp. 2234--2242, 2016.

\bibitem{kawahara2018seven}
J.~Kawahara, S.~Daneshvar, G.~Argenziano, and G.~Hamarneh, ``Seven-point
  checklist and skin lesion classification using multitask multimodal neural
  nets,'' \emph{IEEE Journal of Biomedical and Health Informatics}, vol.~23,
  no.~2, pp. 538--546, 2018.

\bibitem{saida2004significance}
T.~Saida, A.~Miyazaki, S.~Oguchi, Y.~Ishihara, Y.~Yamazaki, S.~Murase,
  S.~Yoshikawa, T.~Tsuchida, Y.~Kawabata, and K.~Tamaki, ``Significance of
  dermoscopic patterns in detecting malignant melanoma on acral volar skin:
  results of a multicenter study in japan,'' \emph{Archives of Dermatology},
  vol. 140, no.~10, pp. 1233--1238, 2004.

\bibitem{iyatomi2008computer}
H.~Iyatomi, H.~Oka, M.~E. Celebi, K.~Ogawa, G.~Argenziano, H.~P. Soyer,
  H.~Koga, T.~Saida, K.~Ohara, and M.~Tanaka, ``Computer-based classification
  of dermoscopy images of melanocytic lesions on acral volar skin,''
  \emph{Journal of Investigative Dermatology}, vol. 128, no.~8, pp. 2049--2054,
  2008.

\bibitem{tan2019efficientnet}
M.~Tan and Q.~V. Le, ``Efficientnet: Rethinking model scaling for convolutional
  neural networks,'' \emph{International Conference on Machine Learning}, pp.
  6105--6114, 2019.

\bibitem{cubuk2020randaugment}
E.~D. Cubuk, B.~Zoph, J.~Shlens, and Q.~V. Le, ``Randaugment: Practical
  automated data augmentation with a reduced search space,'' \emph{IEEE
  Conference on Computer Vision and Pattern Recognition Workshops}, pp.
  702--703, 2020.

\bibitem{huang2016learning}
C.~Huang, Y.~Li, C.~C. Loy, and X.~Tang, ``Learning deep representation for
  imbalanced classification,'' \emph{IEEE Conference on Computer Vision and
  Pattern Recognition}, pp. 5375--5384, 2016.

\bibitem{qian1999momentum}
N.~Qian, ``On the momentum term in gradient descent learning algorithms,''
  \emph{Neural Networks}, vol.~12, no.~1, pp. 145--151, 1999.

\end{thebibliography}
